\documentclass{article}

\usepackage[nonatbib,final]{nips_2016}
\usepackage[utf8]{inputenc} % allow utf-8 input
\usepackage{color}
\usepackage{amsmath}
\usepackage{amssymb}
\usepackage{stmaryrd}
\usepackage{footmisc}
\usepackage{url}
\usepackage{tabularx}
\usepackage{subfigure}
\usepackage{graphicx}
\usepackage{enumerate}
\usepackage{xspace}
\usepackage{mathtools, cuted}
\usepackage{bbm}
\usepackage{multirow,bigdelim}
\usepackage{pifont}
\usepackage{times}
\usepackage{booktabs}
\usepackage{actionable}
\DeclareMathOperator*{\argmax}{arg\,max}
\usepackage{algpseudocode}

\newcommand{\xhdr}[1]{\vspace{1.7mm}\noindent{{\bf #1.}}}
\newcommand{\hide}[1]{}
\newcommand\Fontvi{\fontsize{7}{7.2}\selectfont}

\title{Learning Cost-Effective and Interpretable Regimes \\for Treatment Recommendation}

\author{
  Himabindu Lakkaraju\\
  Department of Computer Science\\
  Stanford University\\
  \texttt{himalv@cs.stanford.edu} \\
  \And
  Cynthia Rudin \\
  Department of Computer Science \\
  Duke University\\
  \texttt{cynthia@cs.duke.edu} \\
}

\begin{document}

\maketitle

\hide{
\begin{abstract}
 Decision makers, such as doctors, make crucial decisions such as recommending treatments to patients on a daily basis. Such decisions typically involve weighing the potential benefits of taking an action against the costs involved. 
In this work, we aim to automate this task of learning \emph{cost-effective, interpretable and actionable treatment regimes}. We formulate this as a problem of learning a decision list -- a sequence of if-then-else rules -- which maps characteristics of subjects (eg., diagnostic test results of patients) to treatments. We propose a novel objective to construct a decision list which maximizes outcomes for the population, and minimizes overall costs. We model the problem of learning such a list as a Markov Decision Process (MDP) and employ a variant of the Upper Confidence Bound for Trees (UCT) strategy which leverages customized checks for pruning the search space effectively.
%develop a customized optimization and search procedure inspired by Monte which significantly reduces the size of the search space. 
Experimental results on real world observational data capturing treatment recommendations for asthma patients demonstrate the effectiveness of our approach.
\end{abstract}
}

\section{Introduction}

%Hima: first few paras in the intro look really good. Thanks Cynthia!
Decision makers, such as doctors, make crucial decisions such as recommending treatments to patients on a daily basis. Such decisions typically involve careful assessment of the subject's condition, analyzing the costs associated with the possible actions, and the nature of the consequent outcomes. Further, there might be costs associated with the assessment of the subject's condition itself (e.g., physical pain endured during medical tests, monetary costs etc.). Decision makers often leverage personal experience to make decisions in these contexts, without considering data, even if massive amounts of it exist. Machine learning models could be of immense help in such scenarios -- but these models would need to consider all three aspects discussed above: predictions of counterfactuals, costs of gathering information, and costs of treatments. Further, these models must be interpretable in order to create any reasonable chance of a human decision maker actually using them. In this work, we address the problem of learning such cost-effective, interpretable treatment regimes from observational data. 

Prior research addresses various aspects of the problem at hand in isolation. For instance, there exists a large body of literature on estimating treatment effects~\cite{d2007estimating,mcgough2009estimating,dorresteijn2011estimating}, recommending optimal treatments~\cite{abulesz1988novel,wallace2014personalizing,fan2016sequential}, and learning intelligible models for prediction~\cite{letham2015interpretable,lakkarajuinterpretable,lou2012intelligible,bien2009classification}. However, an effective solution for the problem at hand should ideally incorporate all of the aforementioned aspects. Furthermore, existing solutions for learning treatment regimes neither account for the costs associated with gathering the required information, nor the treatment costs. The goal of this work is to propose a framework which jointly addresses all of the aforementioned aspects.

We address the problem at hand by formulating it as a task of learning a decision list that maps subject characteristics to treatments such that it: 1) maximizes the expectation of a pre-specified outcome when used to assign treatments to a population of interest 2) minimizes costs associated with assessing subjects' conditions and 3) minimizes costs associated with the treatments themselves. We propose a novel objective function to learn a decision list optimized with respect to the criterion highlighted above. We show that the proposed objective is NP-hard. We then optimize this objective by modeling it as a Markov Decision Process (MDP) and employing a variant of the Upper Confidence Bound for Trees (UCT) strategy which leverages customized checks for pruning the search space effectively.
Our results on a real world dataset comprised of treatment recommendations for asthma patients demonstrate the effectiveness of the proposed solution. 
\vspace{-0.1in}
\section{Our Framework}
First, we formalize the notion of treatment regimes and discuss how to represent them as decision lists. We then propose an objective function for constructing cost-effective treatment regimes.

\xhdr{Input Data and Cost Functions} Consider a dataset $\mathcal{D} = \{(\textbf{x}_1, a_1, y_1),$  $(\textbf{x}_2, a_2, y_2)$ $\cdots$
$(\textbf{x}_N, a_N, y_N) \}$ comprised of $N$ independent and identically distributed observations, each of which corresponds to a \emph{subject} (individual), potentially from an observational study. %Let $\mathcal{X}$ denote the characteristics of all the subjects in $\mathcal{D}$ i.e., $\mathcal{X} = \{\textbf{x}_1, \textbf{x}_2, \cdots \textbf{x}_N\}. 
Let $\textbf{x}_i = \left[ x^{(1)}_i, x^{(2)}_i, \cdots x^{(p)}_i \right] \in \left[\mathcal{V}_1, \mathcal{V}_2, \cdots \mathcal{V}_p\right]$ denote the \emph{characteristics} of subject $i$. $\mathcal{V}_f$ denotes the set of all possible values that can be assumed by a characteristic $f \in \mathcal{F} = \{1, 2, \cdots p\}$. Each characteristic $f \in \mathcal{F}$ can either be a binary, categorical or real valued variable.  In the medical setting, example characteristics include patient's age, BMI, gender, glucose level etc., Let $a_i \in \mathcal{A} = \{1,2,\cdots m\}$ and $y_i \in \mathbbm{R}$ denote the \emph{treatment} assigned to subject $i$ and the corresponding \emph{outcome} respectively. We assume that $y_i$ is defined such that higher values indicate better outcomes. For example, the outcome of a patient can be regarded as a wellness improvement score that indicates the effectiveness of the assigned treatment. 

It can be much more expensive to determine certain subject characteristics compared to others. For instance, a patient's age can be easily retrieved either from previous records or by asking the patient. On the other hand, determining her glucose level requires more comprehensive testing, and is therefore more expensive in terms of monetary costs, time and effort required both from the patient as well as the clinicians. We assume access to functions $d: \mathcal{F} \rightarrow \mathbbm{R}$, and $d': \mathcal{A} \rightarrow \mathbbm{R}$ which return the cost of determining any characteristic in $\mathcal{F}$, and the cost of each treatment $a \in \mathcal{A}$ respectively.

\xhdr{Treatment Regimes}
A treatment regime is a function that takes as input the characteristics of any given subject $\textbf{x}$ and maps them to an appropriate treatment $a 
\in \mathcal{A}$.
% This should be able to be shortened
%Several classes of models ranging from complex, non-linear functions to more interpretable rule-based representations can be used to express such treatment regimes. While more complex functions offer greater flexibility in modeling the data, it is often hard for humans to understand the logic employed by such models. 
We employ \emph{decision lists} to express treatment regimes as they tend to be intelligible. A decision list is an ordered list of rules embedded within an if-then-else structure. A treatment regime expressed as a decision list $\pi$ 
\hide{takes the following form:\\
if $c_1$, then $a_1$ \\
else if $c_2$, then $a_2$ \\ 
$\cdot$ \\ 
$\cdot$ \\ 
else if $c_{L}$, then $a_{L}$\\
else $a_{L+1}$.

The above list } is a sequence of $L+1$ rules $\left[ r_1, r_2, \cdots, r_{L+1}\right]$. 
%(one rule corresponding to each line in the list). 
The last one, $r_{L+1}$, is a default rule which applies to all those subjects who do not satisfy any of the previous $L$ rules. Each rule $r_j$ (except the default rule) is a tuple of the form $(c_j, a_j)$ where $a_j \in \mathcal{A}$, and $c_j$ represents a \emph{pattern} which is a conjunction of one or more predicates. Each predicate takes the form $(f,o,v)$ where $f \in \mathcal{F}$, $o \in \{=, \neq, \leq, \geq, <, > \}$, and $v \in \mathcal{V}_f$ denotes some value $v$ that can be assumed by the characteristic $f$ (Eg.,`Age $\geq$ 40 $\wedge$ Gender$=$Female"). We define an indicator function, $\textit{satisfy}(x_i, c_j)$, which returns a $1$ if $x_i$ satisfies $c_j$ and $0$ otherwise. 

The rules in $\pi$ partition the dataset $\mathcal{D}$ into $L + 1$ groups: $\{\mathcal{R}_1, \mathcal{R}_2 \cdots \mathcal{R}_{L}, \mathcal{R}_{\text{default}}\}$. A group $\mathcal{R}_j$, where $j \in \{1, 2, \cdots L\}$, is comprised of those subjects that satisfy $c_j$ but do not satisfy any of $c_1, c_2, \cdots c_{j-1}$. This can be formally written as:
\begin{equation}\label{eqn:rj} \mathcal{R}_j = \left\{ \textbf{x} \in \left[ \mathcal{V}_1 \cdots \mathcal{V}_p \right]\text{ } | \text{ satisfy}(\textbf{x}, c_j\} \wedge \bigwedge\limits_{t=1}^{j-1} \neg \text{ satisfy}(\textbf{x},c_t) \right\}.
\end{equation}
%Analogously, the group $\mathcal{R}_\text{default}$ contains those subjects that do not satisfy any of the $L$ antecedents in the list i.e., 
%\[ \mathcal{R}_\text{default} = \left\{ \textbf{x} \in \left[ \mathcal{V}_1 \cdots \mathcal{V}_p \right]\text{ } | \bigwedge\limits_{t=1}^{L} \neg \text{ satisfy}(\textbf{x},c_t) \right\}.\]

The treatment assigned to each subject by $\pi$ is determined by the group that he/she belongs to. %For instance, if subject $i$ with characteristics $\mathbf{x}_i$ belongs to group $\mathcal{R}_j$ induced by $\pi$ i.e., $\mathbf{x}_i \in \mathcal{R}_j$, then subject $i$ will be assigned the corresponding treatment $a_j$ under the regime $\pi$. 
More formally,
\begin{equation}\label{eqn:pix}
\pi(\mathbf{x}_i) =  \sum\limits_{l=1}^{L} a_l \text{ } \mathbbm{1}(\textbf{x}_i \in \mathcal{R}_l) + a_\text{default} \text{ } \mathbbm{1}(\textbf{x}_i \in \mathcal{R}_\text{default}) 
\end{equation}
%where $\mathbbm{1}$ denotes an indicator function that returns $1$ if the condition within the brackets evaluates to true and $0$ otherwise. %Thus, $\pi$ returns the treatment assigned to $\mathbf{x}_i$.

Similarly, the cost incurred when we assign a treatment to the subject $i$ (\emph{treatment cost}) according to the regime $\pi$ is given by:
\begin{equation}\label{eqn:phi} \phi(\textbf{x}_i) = d'(\pi(\textbf{x}_i)) \end{equation}
where the function $d'$ (defined previously) takes as input a treatment $a \in \mathcal{A}$ and returns its cost.

We can also define the cost incurred in assessing the condition of a subject $i$ (\emph{assessment cost})  as per the regime $\pi$. Note that a subject $i$ belongs to the group $\mathcal{R}_j$ if and only if the subject does not satisfy the conditions $c_1 \cdots c_{j-1}$, but satisfies the condition $c_j$ (Refer to Eqn. \ref{eqn:rj}).
This implies that the assessment cost incurred for this subject $i$ is the sum of the costs of all the characteristics that appear in $c_1 \cdots c_{j}$. If $\mathcal{N}_l$ denotes the set of all the characteristics that appear in $c_1 \cdots c_{l}$, the assessment cost of the subject $i$ as per the regime $\pi$ can be written as: 

\begin{equation}\label{eqn:psi}
\psi(\textbf{x}_i) = \sum\limits_{l=1}^{L} \left[ \mathbbm{1}(\textbf{x}_i \in \mathcal{R}_l) \times \left( \sum\limits_{e \in \mathcal{N}_l} d(e) \right) \right]. 
\end{equation}

\xhdr{Objective Function}
We first formalize the notions of expected outcome, assessment, and treatment costs of a treatment regime $\pi$ with respect to the dataset $\mathcal{D}$.

The quality of the regime $\pi$ is partly determined by the expected outcome when all the subjects in $\mathcal{D}$ are assigned treatments according to $\pi$. The higher the value of such an expected outcome, the better the quality of the regime $\pi$. There is, however, one caveat to computing the value of this expected outcome -- we only observe the outcome $y_i$ resulting from assigning $\textbf{x}_i$ to $a_i$ in the data $\mathcal{D}$, and not any of the counterfactuals. To address this problem, we compute the expected outcome of a given regime $\pi$ using doubly robust estimation~\cite{lunceford2004stratification}:

\begin{align}\label{eqn:g1}
g_1(\pi) = \frac{1}{N} \sum\limits_{i=1}^{N} \sum\limits_{a \in \mathcal{A}} \left[ \frac{\mathbbm{1}(a_i = a)}{\hat{\omega}(x_i,a)} \{y_i - \hat{y}(x_i, a)\} + \hat{y}(x_i, a) \right]
\mathbbm{1}(\pi(x_i) = a)
\end{align}

$\hat{\omega}(x_i,a)$ denotes the probability that the subject $i$ with characteristics $x_i$ is assigned to treatment $a$ in the data $\mathcal{D}$. $\hat{\omega}$ represents the propensity score model. In practice, we fit a multinomial logistic regression model on $\mathcal{D}$ to learn this function. Our framework does not impose any constraints on the functional form of $\hat{\omega}$. $\hat{y}$ corresponds to the outcome regression model and is learned in our experiments by fitting a linear regression model on $\mathcal{D}$ prior to optimizing for the treatment regimes. 

The assessment cost of a subject $i$ w.r.t. the regime $\pi$ is given in Eqn.~\ref{eqn:psi}. The expected assessment cost across the entire population can be computed as: 
\begin{equation}\label{eqn:g2}
g_2(\pi) = \frac{1}{N} \sum\limits_{i=1}^{N} \psi(\mathbf{x}_i).
\end{equation}
The treatment cost for a subject $i$ who is assigned treatment using regime $\pi$ is given in Eqn.~\ref{eqn:phi}. The expected treatment cost across the entire population can be computed as: 
\begin{equation}\label{eqn:g3}
g_3(\pi) = \frac{1}{N} \sum\limits_{i=1}^{N} \phi(\mathbf{x}_i).
\end{equation}
The smaller the expected assessment and treatment costs of a regime, the more desirable it is in practice. 

Given the observational data $\mathcal{D}$ and a set of all possible combinations of candidate rules $C(\mathcal{L})$, our objective is to maximize $g_1(\pi)$, and minimize $g_2(\pi)$ and $g_3(\pi)$ : 
\begin{equation}\label{eqn:fullobj}
\argmax\limits_{\pi \in C(\mathcal{L}) \times \mathcal{A}} \lambda_1 g_1(\pi) - \lambda_2 g_2(\pi) - \lambda_3 g_3(\pi).
\end{equation}
The $\lambda$'s in Eqn.~\ref{eqn:fullobj} are non-negative weights that scale the relative influence of terms in the objective. 
%In practice, candidate rules are obtained using an association rule mining algorithm such as Apriori~\cite{agrawal1994fast}. 

The above objective function is NP-Hard~\cite{appendix}. We model this objective using a Markov Decision Process and employ a variant of Upper Confidence Bound on Trees (UCT) algorithm which incorporates customized strategies for effectively pruning the search space. A proof for NP-Hardness of Eqn.~\ref{eqn:fullobj} and a detailed optimization procedure for the same is provided in the Appendix~\cite{appendix}.

\section{Experimental Evaluation}
\vspace{-0.1in}
\begin{figure}
\centering
\Fontvi
	\begin{tabular}{|l|}
		\hline \\
		\dsif \attrb{Spiro-Test}\dseq\val{Pos} \dsand \attr{Prev-Asthma}\dseq\val{Yes} \dsand \attr{Cough}\dseq\val{High} \dsthen \classh{C}  %(21.31\%, 91.21\%)
 \\ \\
\dselif \attrb{Spiro-Test}\dseq\val{Pos} \dsand \attr{Prev-Asthma} \dseq \val{No} \dsthen \class{Q} \\ \\
\dselif \attr{Short-Breath} \dseq \val{Yes} \dsand \attr{Gender}\dseq \val{F} \dsand \attr{Age}\dsgeq \val{40} \dsand \attr{Prev-Asthma}\dseq \val{Yes} \dsthen \classh{C} \\ \\
\dselif \attrh{Peak-Flow}\dseq\val{Yes} \dsand \attr{Prev-RespIssue}\dseq\val{No} \dsand \attr{Wheezing} \dseq \val{Yes}, \dsthen \class{Q} \\ \\
\dselif \attr{Chest-Pain}\dseq\val{Yes} \dsand \attr{Prev-RespIssue} \dseq \val{Yes} \dsand \attra{Methacholine} \dseq \val{Pos} \dsthen \classh{C} \\ \\
\dselse \class{Q} \\ \\
		\hline
	\end{tabular}
	\caption{Regime for treatment recommendations for asthma patients output by our framework; \class{Q} refers to quick-relief, and \classh{C} corresponds to controller drugs (\classh{C} is higher cost than \class{Q}); Attributes in \attr{blue} are least expensive.}
	\label{fig:decisionlist}
    \vspace{-0.10in}
\end{figure}

\begin{table}
	\Fontvi
    \centering
	\begin{tabular}{ c c c c c c }
     \toprule
     & Avg. & Avg. & Avg. & Avg. \# of & List \\
	 & Outcome & Assess Cost & Treat Cost & Characs. & Len \\
     \midrule
     \textbf{CITR} & 74.38 & 13.87 & 11.81 & 7.23 & 6 \\
     IPTL & 71.88 & 18.58 & 11.83 & 7.87 & 8 \\
     MCA & 70.32 & 19.53 & 12.01 & 10.23 & -  \\ 
     OWL (Gaussian) & 71.02 & 25 & 12.38 & 16 & - \\
     OWL (Linear) &  71.02 & 25 & 12.38 & 16 & - \\
     \midrule
     Human &  68.32 & - & 12.28 & - & - \\
	 \bottomrule \\
	\end{tabular}
\caption{Results for Treatment Regimes. Our approach: CITR; Baselines: IPTL, MCA, OWL; Human refers to the setting where doctors assigned treatments. }
\label{tab:results}
\end{table}
\begin{table}[t]
\centering
	\scriptsize
    \quad
	\begin{tabular}{ c c } 
     & \textbf{Asthma Dataset}\\
     \midrule
     \# of Data Points & 60048\\
     \midrule
     Characteristics & age, gender, BMI, BP, short breath, temperature, \\
     & cough, chest pain, wheezing, past allergies, asthma history, \\
     & family history, has insurance (cost 1)\\
     & peak flow test (cost = 2) \\
     & spirometry test (cost = 4) \\
     & methacholine test (cost = 6) \\
     \midrule
     Treatments \& Costs &  quick relief (cost = 10) \\
     & controller drugs (cost = 15) \\
     \midrule
     Outcomes \& Scores &  no asthma attack for $\geq$ 4 months (score = 100) \\
     & no asthma attack for 2 months (score = 66) \\
     &  no asthma attack for 1 month (score = 33) \\
     & asthma attack in less than 2 weeks (score = 0) \\
	 \bottomrule
	\end{tabular}
    \vspace{0.1in}
\caption{Summary of dataset.}
\label{tab:datasets}
\vspace{-0.2in}
\end{table}
\normalsize

Here, we discuss the detailed experimental evaluation of our framework. 
First we analyze the outcomes obtained and costs incurred 
when recommending treatments using our approach. Then, we qualitatively analyze the treatment regime produced by our framework.

\xhdr{Dataset} Our dataset (see Table~\ref{tab:datasets}) captures details of about 60K \textbf{asthma} patients~\cite{lakkarajuinterpretable}. For each of these 60K patients, various attributes such as demographics, symptoms, past health history (cost = 1), test results for peak-flow (cost = 2), spirometry (cost = 4), methacholine (cost = 6) have been recorded. Each patient in the dataset was prescribed either quick relief medications (cost = 10) or long term controller drugs (cost = 15). Further, the outcomes in the form of time to the next asthma attack (after the treatment began) were recorded. The longer this interval, the better the outcome, and the higher the outcome score. 

\xhdr{Baselines} We compared our framework to the following state-of-the-art treatment recommendation approaches: 1) Outcome Weighted Learning (OWL)~\cite{zhao2012estimating} 2) Modified Covariate Approach (MCA)~\cite{tian2014simple} 3) Interpretable and Parsimonious Treatment Regime Learning (IPTL)~\cite{BIOM:BIOM12354}. While none of these approaches explicitly account for treatment costs or costs required for gathering the subject characteristics, MCA and IPTL minimize the number of characteristics/covariates required for deciding the treatment of any given subject. OWL, on the other hand, often uses all the characteristics available in the data when assigning treatments. 

\xhdr{Quantitative Analysis} We analyzed the performance of our approach CITR (Cost-effective, Interpretable Treatment Regimes) w.r.t. various metrics such as: average outcome obtained (Avg. Outcome), average assessment and treatment costs (Avg. Assess Cost, Avg. Treat Cost), average no. of characteristics (Avg. \# of Characs.) used to determine treatment of any given patient, and number of rules in the rule list (List Len).  These results are shown in Table~\ref{tab:results}.  It can be seen that the treatment regimes produced by our approach results in better average outcomes with lower costs.  It is also interesting that our approach produces more concise lists with fewer rules compared to the baselines. While the treatment costs of all the baselines are similar, there is some variation in the average assessment costs and the outcomes. IPTL turns out to be the best performing baseline in terms of the average outcome, average assessment costs, and average number of characteristics. The last line of Table \ref{tab:results} shows the average outcomes and the average treatment costs computed empirically on the observational data. %These statistics correspond to those of the human experts and it is interesting that the regimes learned by algorithmic approaches perform better than human experts on both of the datasets.

\xhdr{Qualitative Analysis} The treatment regime produced by our approach on the asthma dataset is shown in Figure~\ref{fig:decisionlist}. It can be seen that methacholine test, which is more expensive, appears at the end of the regime. This ensures that only a small fraction of the population (8.23\%) is burdened by its cost. Furthermore, it turns out that though the spirometry test is slightly more expensive than patient demographics and symptoms, it would be harder to determine the treatment for a patient without this test. This aligns with research on asthma treatment recommendations~\cite{pereira2015breath,boulet2015benefits}. Furthermore, it is interesting to note that the regime not only accounts for test results on spirometry and peak flow but also assesses if the patient has a previous history of asthma or respiratory issues. %If the test results are positive and the patient has no previous history of asthma or respiratory disorders, then the patient is recommended quick relief drugs. On the other hand, if the test results are positive and the patient suffered previous asthma or respiratory issues, then controller drugs are recommended.

\section{Conclusions}
In this work, we proposed a framework for learning cost-effective, interpretable treatment regimes from observational data. To the best of our knowledge, this is the first solution to the problem at hand that addresses all of the following aspects: 1) maximizing the outcomes 2) minimizing treatment costs, and costs associated with gathering information required to determine the treatment 3) expressing regimes using an interpretable model. We modeled the problem of learning a treatment regime as a MDP and employed a variant of UCT which prunes the search space using customized checks. We demonstrated the effectiveness of our framework on real world data from health care domain.

\small
\bibliographystyle{abbrv} 
\bibliography{actionable}
\end{document}